\documentclass[letterpaper]{article} 
\usepackage[preprint]{aaai2027}  
\usepackage[hyphens]{url}  
\usepackage{graphicx} 
\usepackage{natbib}  
\usepackage{caption} 
\usepackage{algorithm}
\usepackage{algorithmic}
\usepackage{booktabs}
\usepackage{amsmath,amssymb,amsthm}
\usepackage{array}

\newtheorem{definition}{Definition}
\newtheorem{proposition}{Proposition}
\newtheorem{theorem}{Theorem}
\newtheorem{lemma}{Lemma}

\newcommand{\racl}{\textsc{RACL}}
\newcommand{\fvr}{\mathrm{FVR}}
\newcommand{\edr}{\mathrm{EDR}}

\newcommand{\argmax}{\mathop{\mathrm{argmax}}}
\newcolumntype{L}[1]{>{\raggedright\arraybackslash}p{#1}}

\title{Repair Before Veto: Repair-Augmented Constraint Learning for Contextual Decisions}
\author{Yifan Wang}
\affiliations{
Department of Mechanical Engineering, McGill University, QC, H3A 2T7, Canada\\
yifan.wang18@mail.mcgill.ca
}

\begin{document}

\maketitle

\begin{abstract}
Hard constraints are usually treated as terminal vetoes: once a candidate
violates a requirement, the learned rule rejects it and any repair is handled
outside the decision semantics. This misses a common deployed regime in which
the system already knows a finite menu of modifications, such as adding a ticket
option, changing a configuration, or requesting an available service upgrade.
Existing constraint-learning, soft-relaxation, and recourse methods address
nearby problems, but they do not learn whether an option should be repaired
before being vetoed. We introduce Repair-Augmented Constraint Learning
(\racl), a contextual decision framework that lifts known repair operators into
the classifier semantics. A candidate is accepted when an affordable repair
makes it feasible and preferred enough; otherwise the system returns a
structured rejection credit and, when applicable, a repair plan. This
repair-before-veto view strictly generalizes no-repair HASSLE-style semantics,
reveals an irreducible false-veto gap for terminal-veto rules, separates
binary-label non-identifiability from decision-rule learnability, and gives
capacity and calibration bounds for the observed-feasibility shared-weight
setting. Across controlled and DB1B-derived benchmarks, \racl{} recovers the
intended credit and repair structure. On the hardest raw-data-derived tier,
validation-selected \racl{} reduces false vetoes to $10/4039$ (FVR $0.0025$),
versus about $1064/4039$ for the strongest repair-search black-box baseline,
while making the FVR/EDR trade-off explicit.
\end{abstract}

\section{Introduction}

Many AI decision pipelines are built around a simple division of labor:
constraints veto, preferences rank. This design is effective when a candidate is
immutable, but it is brittle when the system can act. In travel, retail,
configuration, scheduling, and service recommendation, a platform often knows
how to modify a candidate before showing it to a user: add a checked bag, switch
to a refundable fare, attach a missing accessory, or adjust an available
resource. A terminal veto discards such candidates even when a low-cost repair
would make them feasible and attractive. The error is not merely poor ranking;
it is a false veto.

This failure mode is not covered by existing formulations. Constraint and
preference learning, including HASSLE-style contextual optimization, learns when
a candidate is feasible and good in its current context
\cite{Kumar2020HASSLE,Kumar2023ContextualMaxSAT}. Soft-constraint and penalty
methods trade feasibility against objective value, but do not return an
executable repair that restores hard feasibility. Recourse and counterfactual
methods search for changes after a model has already rejected a candidate. The
missing decision problem is earlier and sharper: before vetoing an option,
should the system apply one of its known affordable repairs, and if not, why not?

We call this the repair-before-veto problem. The input is a candidate, a
context, and a known repair ontology: a finite set of operators with observable
effects, applicability conditions, and costs. The output includes a binary
decision, a credit category explaining the rejection or acceptance, and a repair
plan when one is selected. This setting is narrower than arbitrary repair
discovery, but it matches many deployed systems: available modifications are
known, while the contextual value of those modifications must be learned.

\begin{figure*}[t]
\centering
\includegraphics[width=.95\textwidth]{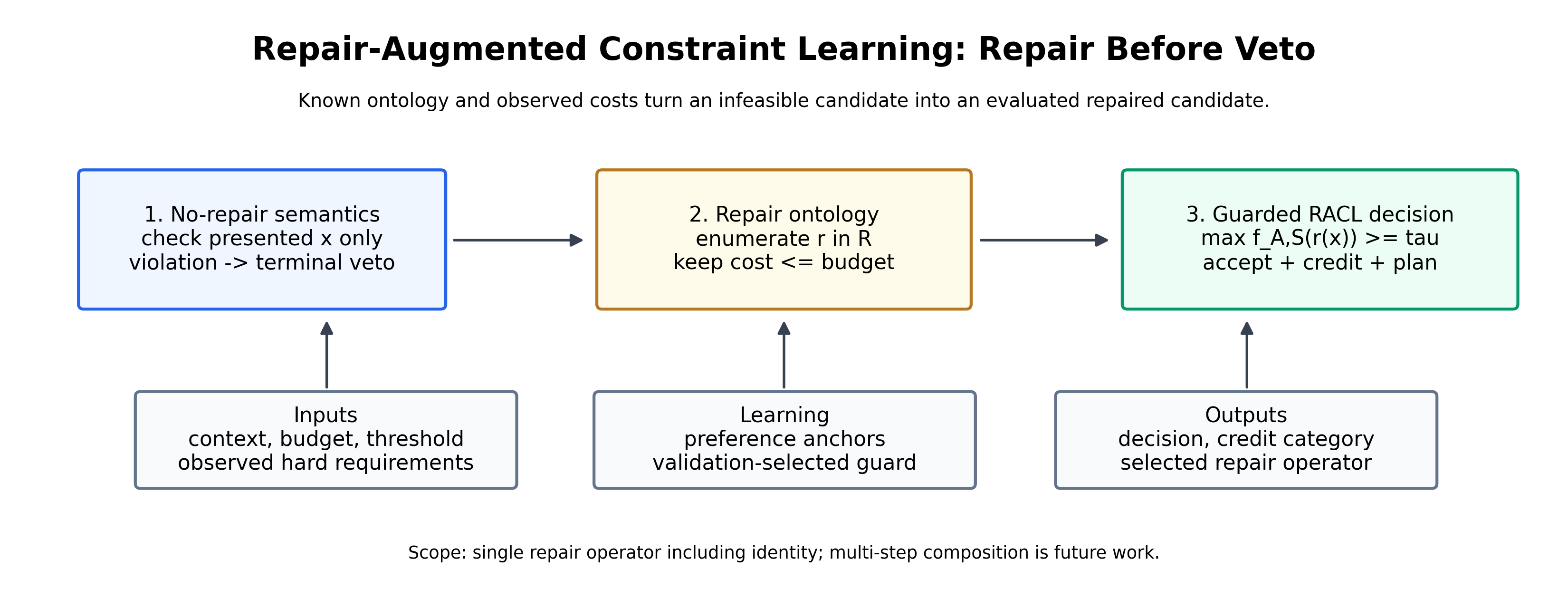}
\caption{Overview of \racl{}. A no-repair rule makes violation terminal; \racl{}
instead enumerates a known repair ontology, checks cost against the contextual
budget, and uses a guarded post-repair score to return a decision, credit
category, and selected repair operator.}
\label{fig:overview}
\end{figure*}

\racl{} lifts repair search into the semantics of constraint learning. A
no-repair rule accepts only if the presented candidate is feasible and preferred;
\racl{} accepts if an affordable repaired candidate is feasible and preferred.
Figure~\ref{fig:overview} summarizes the pipeline. The semantic change alters
credit assignment: negative labels may arise from impossible repair, excessive
repair cost, or insufficient post-repair value, and collapsing these cases into
one rejection cannot support reliable repair-aware decisions. \racl{} is
therefore neither soft relaxation nor post-hoc recourse. It is a
decision-learning semantics that makes repair a first-class action.

This paper makes four contributions.
\begin{itemize}
\item We introduce a repair-augmented contextual decision semantics that
combines executable repair operators, budget-aware value, credit categories,
and repair-plan output.
\item We identify the core statistical obstacle: terminal-veto semantics incur a
conditional false-veto gap, while binary labels alone cannot identify
repairability, cost, or preference scale without contextual budget variation and
anchors.
\item We give pseudo-dimension and sample-complexity bounds for the
observed-feasibility \racl{} class, including a $\widetilde O(p\log K)$
shared-weight bound and a calibration-error bound for guarded decisions.
\item We provide a four-tier evaluation against label-only, credit-supervised,
and repair-search baselines, showing when \racl{} closes the false-veto gap and
where the raw-data-derived setting induces a transparent FVR/EDR trade-off.
\end{itemize}

\section{Related Work}

\paragraph{Constraint and preference learning.}
\racl{} is closest to contextual constraint learning and preference-aware
combinatorial optimization. HASSLE learns MAX-SAT-style feasibility and
preference models from contextual examples \cite{Kumar2020HASSLE,Kumar2023ContextualMaxSAT};
related symbolic learning methods include SMT-based constraint learning,
classical constraint acquisition, and constraint synthesis from examples
\cite{Kolb2018SMT,Bessiere2013QuAcq,Bessiere2017ConstraintAcquisition}. These methods
learn or acquire constraints, but they typically treat infeasibility as a final
state rather than as a state from which an affordable transformation may recover
the candidate. \racl{} also connects to constraint programming, SAT, and answer
set solving \cite{Rossi2006CPHandbook,Biere2009SATHandbook,Gebser2012ASP}, but
the present contribution is a decision-learning semantics rather than a new
complete solver.

\paragraph{Relaxation and over-constrained problems.}
Classical partial CSP, soft constraints, and semiring-based CSPs address
over-constrained problems by relaxing or weighting constraints
\cite{Freuder1992PartialCSP,Bistarelli1997SemiringCSP,Meseguer2006SoftConstraints}.
\racl{} is complementary: constraints are not merely softened. Instead, a
candidate may be transformed by a known repair operator, after which feasibility,
cost, and post-repair preference are evaluated jointly. This candidate-repair
view is essential for returning an executable plan and a credit category.

\paragraph{Structured learning and symbolic supervision.}
Structured prediction and symbolic regularization provide tools for learning
with output dependencies and declarative knowledge
\cite{Tsochantaridis2004StructuredSVM,Xu2018SemanticLoss}.
\racl{} differs in the role of the structure. Repair operators are not merely
constraints on predictions; they define possible transformations of candidates,
and decisions are made by maximizing value over the affordable transformed set.

\paragraph{Repair, recourse, and recommender systems.}
Counterfactual explanations and algorithmic recourse seek actionable changes
that alter a fixed model's output
\cite{Wachter2018Counterfactual,Ustun2019Recourse,Karimi2021RecourseInterventions}.
\racl{} uses related language but solves a different problem: repairs are part
of the decision rule before the veto is issued, not only an explanation after a
black-box decision. Constraint-based recommender systems also reason with
requirements, repair suggestions, and preference models
\cite{Felfernig2008ConstraintRecommenders,Felfernig2013RepairScoringRules}.
\racl{} contributes a statistical learning formulation and evaluation protocol
for this repair-aware decision setting. Preference representations such as
CP-nets and broader preference-learning methods provide complementary tools for
user preference modeling \cite{Boutilier2004CPNets,Domshlak2011PreferencesAI}.

\paragraph{Calibration and decision metrics.}
Because \racl{} compares learned post-repair values with contextual thresholds,
calibration matters. We use validation-only guard selection and report both
ranking and decision metrics, drawing on standard calibration and ROC analysis
\cite{Platt1999ProbabilisticOutputs,Zadrozny2002Calibration,Guo2017Calibration,Fawcett2006ROC,Davis2006PRROC}.

\section{Problem Setting}

A candidate $x$ is evaluated in a context $\psi$. The context contains a budget
$b(\psi)$, an acceptability threshold $\tau(\psi)$, and possibly user or persona
features. Let $\phi_\psi$ denote observed hard requirements, and let
$f(x)=\langle w,g(x)\rangle$ be a preference score over candidate features. A
finite repair ontology $\mathcal R=\{r_1,\ldots,r_K\}$ contains the identity
repair and structured operators that transform candidates. A repair has an
observable cost $\rho(r,x,\psi)$ and may change feasibility and value.

\begin{definition}[Budget-only \racl{} decision rule]
The admissible repair set is
\[
\mathcal A(x,\psi)=\{r\in\mathcal R: r(x)\models\phi_\psi,\,
\rho(r,x,\psi)\le b(\psi)\}.
\]
The repair-aware value is
\[
V^0(x,\psi)=\max_{r\in\mathcal A(x,\psi)} f(r(x)),
\qquad \max_{\emptyset}=-\infty.
\]
\racl{} accepts if and only if
\[
y(x,\psi)=\mathbf 1[V^0(x,\psi)\ge \tau(\psi)].
\]
\end{definition}

The main model uses repair cost as an affordability gate. A cost-tradeoff rule
$\max_r f(r(x))-\lambda \rho(r,x,\psi)$ is a natural extension, but it is not
the claim evaluated here.

\paragraph{Credit categories and plans.}
\racl{} uses one priority-ordered taxonomy: accepted-already-good,
accepted-repairable-good, rejected-non-repairable,
rejected-repairable-over-budget, rejected-feasible-suboptimal, and
rejected-repairable-suboptimal. The evaluation buckets merge the two structural
rejection categories as structure/cost, merge the two value-failure categories as
preference-dependent, and keep accepted-repairable-good separate from other
accepted cases. The present model selects one repair operator, including the
identity; multi-step repair composition with accumulated costs is left to future
work. A repair plan is therefore the selected operator returned on acceptance.

\paragraph{Metrics.}
Let $\mathcal G_{\mathrm{repair}}$ be the repairable-good region later defined in
Proposition~2, and let $\hat y_M$ be model $M$'s decision. We report
\[
\fvr(M)=
\frac{|\{(x,\psi)\in\mathcal G_{\mathrm{repair}}:\hat y_M(x,\psi)=0\}|}
{|\mathcal G_{\mathrm{repair}}|}.
\]
For decision regret, let $V^\star(C,\psi)$ be the oracle value of the best option
in a presented choice set $C$, and let $V^M(C,\psi)$ be the ground-truth value of
the option selected by model $M$. We report
\[
\edr(M)=\mathbb E_{(C,\psi)}[V^\star(C,\psi)-V^M(C,\psi)].
\]
The value scale is the normalized persona preference score used by the benchmark
generator. Credit accuracy evaluates the rejection credit category; repair-plan
accuracy evaluates exact plan recovery for repairable candidates. AUROC is
reported as a secondary generic ranking metric. This metric choice reflects the
target application: losing a repairable-good option is operationally different
from a conventional classification error. We therefore also consider an
application-level cost $c_{\mathrm{fv}}\fvr+c_{\mathrm{reg}}\edr$ in the
supplement: false vetoes remove actionable feasible outcomes, while regret
measures suboptimal choices among outcomes that remain available.

\section{Repair-Augmented Learning}

The implementation follows Algorithm~\ref{alg:racl}. The learner uses feasible
non-repaired anchors to fit per-persona preference scores, then enumerates the
known repair ontology at decision time. In the raw DB1B-derived setting there are
no clean zero-budget contexts, so feasible candidates serve as preference
anchors; this is a documented limitation and motivates the calibration analysis.

\begin{algorithm}[tb]
\caption{\racl{} Decision with Validation-Selected Guard}
\label{alg:racl}
\textbf{Input}: candidate $x$, context $\psi$, repair ontology $\mathcal R$,
learned preference score $\hat f$, guard parameters $(A,S)$\\
\textbf{Output}: decision, credit category, optional selected repair
\begin{algorithmic}[1]
\STATE $\mathcal A \leftarrow \emptyset$
\FORALL{$r\in\mathcal R$}
  \IF{$r(x)\models\phi_\psi$ and $\rho(r,x,\psi)\le b(\psi)$}
     \STATE add $r$ to $\mathcal A$
  \ENDIF
\ENDFOR
\IF{$\mathcal A=\emptyset$}
  \STATE \textbf{return} reject, non-repairable or over-budget credit, no plan
\ENDIF
\STATE $r^\star\leftarrow \argmax_{r\in\mathcal A}\hat f(r(x))$
\IF{$\hat f_{A,S}(r^\star(x))\ge \tau(\psi)$}
  \STATE \textbf{return} accept, already-good or repairable-good credit, $r^\star$
\ELSE
  \STATE \textbf{return} reject, feasible- or repairable-suboptimal credit, no plan
\ENDIF
\end{algorithmic}
\end{algorithm}

\paragraph{Validation-selected guard.}
Raw candidate distributions can make affine calibration unstable near the
acceptability threshold. Let a per-persona logistic anchor model have coefficient
vector $u$, intercept $d$, and coefficient $\beta_\tau$ on the feature
$-\tau$. For guard $(A,S)$, if $\beta_\tau\ge 1/A$ we first set
$\tilde f(x)=(u^\top g(x)+d)/\beta_\tau$. If the anchor-score span
$R=\max_i\tilde f(x_i)-\min_i\tilde f(x_i)$ exceeds $S$ times the observed
threshold span $T=\max_i\tau_i-\min_i\tau_i$, we replace it by
\[
\hat f_{A,S}(x)=\frac{2T}{R}\{\tilde f(x)-m\}+\bar\tau,
\]
where $m$ is the anchor-score midpoint and $\bar\tau$ is the threshold midpoint.
When $\beta_\tau<1/A$, we use the same span guard after a normalized-score
fallback. Thus $A$ caps unstable affine scaling and $S$ caps score span relative
to the threshold span. We use the predeclared grid
$A\in\{5,10,20,50\}$ and $S\in\{5,10,20\}$, selecting on validation data only by
FVR first, then EDR and AUROC. The selected model is refit on the training prefix
and evaluated once on test data.

\section{Theoretical Properties}

The following statements summarize the formal evidence used in the paper. The
main text gives the core proof ideas because they are central to the model's
scope. Full proofs, calibration diagnostics, and additional tables are provided
in the supplementary technical appendix.

\begin{proposition}[Strict generalization]
If $\mathcal R=\{\mathrm{id}\}$, then \racl{} reduces to the no-repair
feasibility-plus-preference rule $y=1$ iff $x\models\phi_\psi$ and
$f(x)\ge\tau(\psi)$. There exist fixed $(\phi,f)$ instances in which the same
infeasible $x$ is accepted in a high-budget context and rejected in a low-budget
context; no no-repair hypothesis over the same $(\phi,f)$ can represent this
behavior because its decision on an infeasible $x$ is budget-independent.
\end{proposition}

\begin{proposition}[No-repair false-veto gap]
Let
\[
\begin{aligned}
\mathcal G_{\mathrm{repair}}=\{(x,\psi):\;&x\not\models\phi_\psi,\,
\exists r\in\mathcal R,\\
&\rho(r,x,\psi)\le b(\psi),\,
f(r(x))\ge\tau(\psi)\}.
\end{aligned}
\]
Any no-repair rule whose acceptance requires $x\models\phi_\psi$ rejects every
example in $\mathcal G_{\mathrm{repair}}$. Its conditional FVR on this region is
therefore 1; this is a conditional repairable-good error statement, not a global
error-rate claim.
\end{proposition}

\begin{proposition}[Binary-only non-identifiability]
Binary labels alone do not identify the latent repair triple
$(\phi,\rho,f)$. A repairable-over-budget violation and a non-repairable
violation induce identical labels at a fixed budget. Moreover,
$(f,\tau)$ and $(af+c,a\tau+c)$ induce identical labels for any $a>0$.
Contextual budget variation and feasible anchors are therefore necessary to
identify repair-cost intervals and preference scale. This identifiability
statement concerns recovering $(\phi,\rho,f)$; the learnability result below
concerns prediction of the induced decision rule.
\end{proposition}

\begin{theorem}[Capacity and learnability]
Assume the hard requirements $\phi_\psi$ and repair costs are observed, as in
the experiments, $|\mathcal R|=K$, and $f$ is linear with $p$ parameters. Under
the shared-weight regime in which the same $w\in\mathbb R^p$ scores all repaired
candidates, the budget-only \racl{} class has pseudo-dimension
$\widetilde O(p\log K)$. ERM is agnostically PAC learnable with sample complexity
$O((\mathrm{Pdim}+\log(1/\delta))/\varepsilon^2)$.
\end{theorem}

The bound follows standard pseudo-dimension and uniform-convergence arguments
\cite{Vapnik1998SLT,Anthony1999NNLearning,ShalevShwartz2014UML}. With observed
feasibility, each example contributes at most $K$ repaired linear scores and one
threshold comparison; the finite max over repairs adds only logarithmic
combinatorial factors. If feasibility were itself learned from a hypothesis
class, an additional growth-function term for that class would enter; that is
not the regime evaluated here.

\begin{lemma}[Margin-bounded decision error under calibrated scores]
If an affine-calibrated score $\hat f_{\mathrm{cal}}$ has
$\sup_x |\hat f_{\mathrm{cal}}(x)-f^\star(x)|\le\Delta$ on the support, then for
any threshold $\tau$,
\[
\Pr[\mathbf 1[\hat f_{\mathrm{cal}}(x)\ge\tau]\ne
\mathbf 1[f^\star(x)\ge\tau]]
\le
\Pr[|f^\star(x)-\tau|\le\Delta].
\]
\end{lemma}

Thus residual false vetoes concentrate near the acceptability boundary. This
matches the raw-data-derived setting, where feasible anchors are noisier than
zero-budget anchors and the candidate distribution has substantial near-threshold
mass.

\section{Experiments}

\paragraph{Benchmarks.}
We evaluate four tiers. Synthetic-MAXSAT is a controlled Boolean repair task.
Expedia-schema and DB1B-schema are schema-calibrated semi-synthetic tasks based
on public hotel and ticket-field schemas. The hardest tier is DB1B-derived:
it uses real NBER/BTS DB1B ticket records from 2016Q3 for fare, route, carrier,
airport, distance, passenger, and ticket-type distributions \cite{BTSDB1B}, then
injects ancillary repair attributes, contexts, labels, and credit categories.
The Expedia-schema tier uses the public Kaggle Expedia Hotel Recommendations
schema \cite{KaggleExpedia}. We state this up front because the raw tier is a
real-distribution stress test, not a fully natural repair-choice dataset: DB1B
does not contain baggage, seat-selection, refundability, safe-connection,
wrong-date, or user repair-decision fields.

\paragraph{Baselines.}
HASSLE-style NoRepair learns the no-repair feasibility-plus-preference rule.
SoftPenalty-LearnedFeat fits a soft feasibility penalty. BlackBox is a label-only
gradient boosting classifier \cite{Friedman2001GradientBoosting,Pedregosa2011ScikitLearn}.
BlackBox+CreditHead receives explicit credit labels. BlackBox+RepairSearch is
the strongest repair-aware baseline: it receives the same repair ontology and
repair search space as \racl{}, but uses a black-box label model to score repaired
candidates. Oracle-RACL uses ground-truth repair and preference information.
All learned baselines receive the same candidate/context features available to
\racl{}, including observable violation-count features; label-only baselines do
not receive credit labels or repair search.
BlackBox+RepairSearch is intentionally strong: it receives exactly the same
repair ontology and search space as \racl{}, so any remaining difference is a
decision-rule and calibration difference rather than ontology access.

\paragraph{Protocol.}
The raw optimized protocol uses 50,000 training rows and 15,000 test rows. Guard
selection subdivides the training prefix into 45,000 calibration-training rows
and 5,000 validation rows. Multi-seed model runs use seeds $\{1,3,5\}$; guard
stability repeats validation splitting with seeds $\{11,17,23,31,43\}$. Test
labels are never used to select the guard.

\begin{table*}[t]
\centering
\small
\setlength{\tabcolsep}{3pt}
\begin{tabular}{L{.13\textwidth}L{.16\textwidth}L{.16\textwidth}L{.16\textwidth}L{.20\textwidth}}
\toprule
Tier & HASSLE-style NoRepair & BlackBox & \racl{} operating point & Main observation \\
\midrule
Synthetic-MAXSAT &
FVR 1.000, EDR 2.111, Cr 0.210 &
FVR 0.000, EDR 1.658, Cr 0.210 &
FVR 0.000, EDR 0.020, Cr 1.000, Plan 1.000 &
RACL removes false vetoes and recovers structure; BlackBox fits labels but not credit. \\
Expedia-schema &
FVR 1.000, EDR 0.063, Cr 0.589 &
FVR 0.032, EDR 0.060, Cr 0.589 &
FVR 0.000, EDR 0.003, Cr 1.000, Plan 1.000 &
Controlled semi-synthetic tier shows clean repair-structured dominance. \\
DB1B-schema &
FVR 1.000, EDR 0.097, Cr 0.441 &
FVR 0.005, EDR 0.195, Cr 0.441 &
FVR 0.000, EDR 0.000, Cr 1.000, Plan 1.000 &
Schema-calibrated ticket tier preserves the predicted no-repair gap. \\
DB1B-dist. &
FVR 1.000, EDR 0.159, Cr 0.416 &
FVR 0.113, EDR 0.055, Cr 0.416 &
FVR 0.0025, EDR 0.056, Cr 1.000, Plan 1.000 &
Real-distribution tier exposes an FVR/EDR trade-off rather than uniform dominance. \\
\bottomrule
\end{tabular}
\caption{Multi-tier summary. The first three tiers validate the theoretical
prediction in controlled settings: no-repair semantics incur the false-veto gap,
while \racl{} recovers repair-structured credit/plans. The raw DB1B-derived tier
is harder and is analyzed in detail in Table~\ref{tab:raw-main}. Cr denotes
CreditAcc.}
\label{tab:multi-tier}
\end{table*}

\begin{figure}[t]
\centering
\includegraphics[width=\columnwidth]{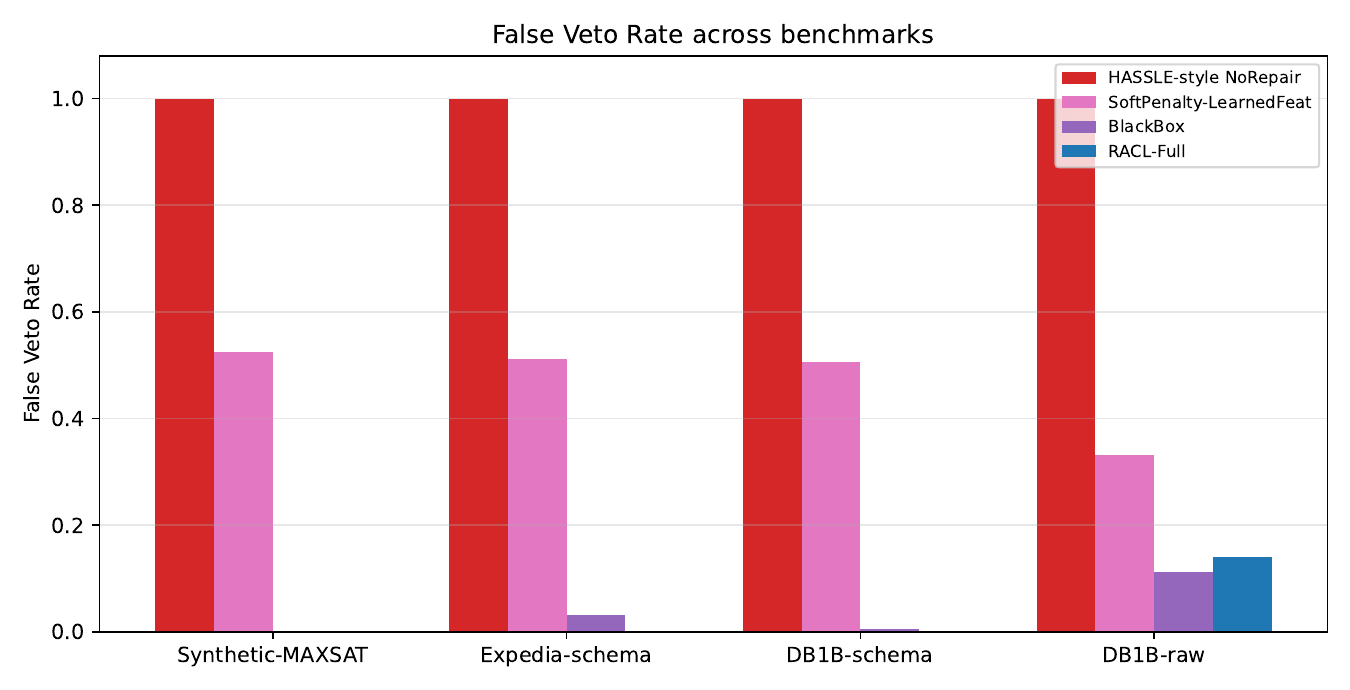}
\caption{False-veto rate across benchmark tiers. The no-repair baseline has
structural FVR 1.0 on repairable-good candidates, while \racl{} removes that gap
in controlled/schema settings and uses validation-selected calibration in the
raw tier.}
\label{fig:fvr}
\end{figure}

\begin{table*}[t]
\centering
\small
\begin{tabular}{L{.18\textwidth}L{.11\textwidth}L{.12\textwidth}L{.14\textwidth}cccc}
\toprule
Model & Repair & Credit labels & FVR (false vetoes / 4039; 95\% CI) & EDR & CreditAcc & PlanAcc & AUROC \\
\midrule
HASSLE-style NoRepair & No & No & 1.0000 (4039; [1.0000,1.0000]) & 0.1591 & 0.4162 & -- & 0.4276 \\
SoftPenalty & No search & No & 0.3325 (1343; [0.3195,0.3470]) & 0.0870 & 0.4162 & -- & 0.8694 \\
BlackBox & No search & No & 0.1129 (456; [0.1039,0.1236]) & 0.0549 & 0.4162 & -- & \textbf{0.9828} \\
BlackBox+RepairSearch & Yes & No & 0.2633 ($\approx$1064; [0.250,0.277]) & \textbf{0.0296} & 1.0000 & 1.0000 & 0.9787 \\
BlackBox+CreditHead & No search & Yes & 0.1129 (456; [0.1035,0.1235]) & 0.0549 & 0.9671 & -- & \textbf{0.9828} \\
\racl-Full & Yes & No & 0.1399 (565; [0.1281,0.1510]) & 0.0540 & 1.0000 & 1.0000 & 0.9632 \\
\racl-ValidatedGuard & Yes & No & \textbf{0.0025 (10; [0.0013,0.0046])} & 0.0563 & \textbf{1.0000} & \textbf{1.0000} & 0.9252 \\
Oracle-\racl & Yes & n/a & 0.0000 (0; [0.0000,0.0010]) & 0.0000 & 1.0000 & 1.0000 & 0.9978 \\
\bottomrule
\end{tabular}
\caption{Raw DB1B-derived semi-synthetic results. \racl-ValidatedGuard is the
best non-oracle model on false-veto rate while preserving perfect structured
credit and repair-plan accuracy. BlackBox+RepairSearch has lower EDR, so the raw
result is an FVR/EDR trade-off rather than universal dominance.}
\label{tab:raw-main}
\end{table*}

\paragraph{Main results.}
Table~\ref{tab:multi-tier} shows the broad pattern. In the controlled and
schema-calibrated tiers, \racl{} removes the no-repair false-veto gap, achieves
near-zero EDR, and recovers credit/plans without credit supervision. The raw
DB1B-derived tier is intentionally harder. Table~\ref{tab:raw-main} shows that
the no-repair baseline has FVR 1.0 because every repairable-good infeasible
candidate is rejected.\footnote{The sub-0.5 AUROC for NoRepair reflects this
deterministic rejection on $\mathcal G_{\mathrm{repair}}$; ranking metrics are
reported only for completeness.} SoftPenalty lowers FVR relative to NoRepair but
remains far from the low-FVR region. Label-only BlackBox reduces FVR to 0.1129
but does not recover repair credit. BlackBox+CreditHead improves credit using explicit credit
supervision but does not output repair plans. BlackBox+RepairSearch recovers
credit and plans when given the same ontology as \racl{}, and it obtains the best
non-oracle EDR. Its FVR, however, is 0.2633. \racl-ValidatedGuard achieves FVR
0.0025: 10 false vetoes among 4039 repairable-good test candidates. The FVR
denominator is exactly $|\mathcal G_{\mathrm{repair}}|$ on the raw test set.
Pushing FVR this low slightly raises EDR relative to \racl-Full; this is the
intended operating-point shift, not a hidden regression. A cost sweep in the
supplement shows that \racl{} is preferred to BlackBox+RepairSearch whenever
$c_{\mathrm{fv}}/c_{\mathrm{reg}}>0.102$.

\begin{figure}[t]
\centering
\includegraphics[width=\columnwidth]{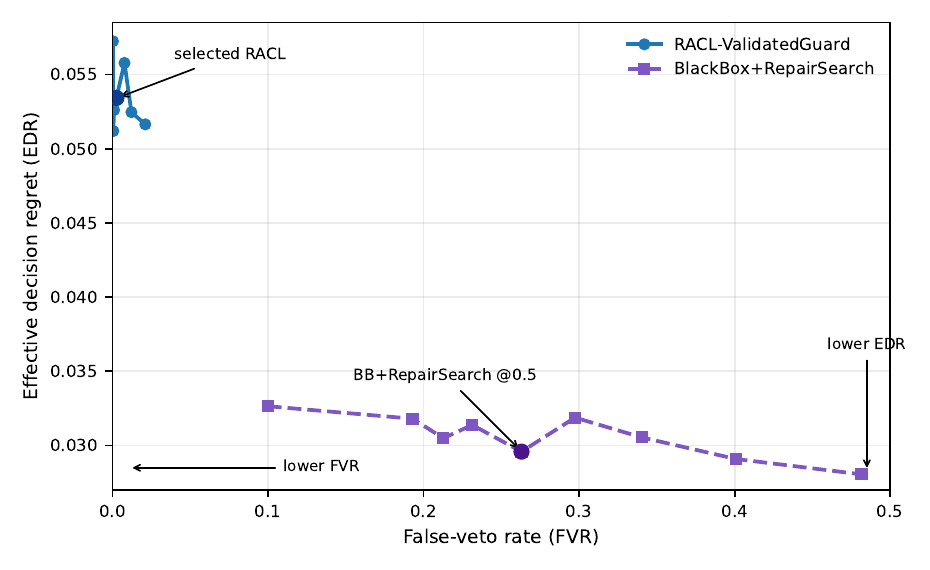}
\caption{Raw DB1B-derived operating sweep. \racl{} thresholds vary the margin
around $\tau$; BlackBox+RepairSearch thresholds vary the repair-search acceptance
probability. Lower-left is preferred. The repair-search black box occupies a
lower-EDR, high-FVR region, while \racl{} targets low FVR.}
\label{fig:sweep}
\end{figure}

\begin{table}[t]
\centering
\small
\begin{tabular}{lccccc}
\toprule
Split & $A_{\max}$ & $S$ & Val FVR & Test FVR & Test AUROC \\
\midrule
11 & 5 & 5 & 0.0036 & 0.0025 & 0.9252 \\
17 & 5 & 5 & 0.0021 & 0.0025 & 0.9252 \\
23 & 5 & 5 & 0.0021 & 0.0025 & 0.9252 \\
31 & 5 & 5 & 0.0093 & 0.0025 & 0.9252 \\
43 & 5 & 5 & 0.0293 & 0.0025 & 0.9252 \\
\bottomrule
\end{tabular}
\caption{Validation-guard stability on raw DB1B-derived data. Only the
guard-selection split varies across rows; because all five splits select
$(A_{\max},S)=(5,5)$ and the final model is refit on the identical training
prefix, the test evaluation is identical by construction, demonstrating guard
stability rather than test tuning.}
\label{tab:guard}
\end{table}

\paragraph{Validation stability and ablations.}
Table~\ref{tab:guard} shows that guard selection is not a lucky split: all five
stratified validation splits select the same operating point. The selected guard
optimizes validation FVR first, then EDR; it does not optimize AUROC. The AUROC
drop from \racl-Full (0.9632) to \racl-ValidatedGuard (0.9252) is therefore an
explicit FVR/AUROC trade-off, and we retain \racl-Full as the
ranking-conservative operating point. Guard-grid sensitivity shows a broad
low-FVR basin around the selected setting rather than a test-tuned knife-edge.
Stress tests further show that \racl{} is robust to moderate cost noise and
calibration noise, but degrades sharply when many repair templates are missing.
This supports the intended scope: \racl{} is a method for decision learning
under a reasonably complete known repair ontology.

\paragraph{Credit dependency.}
The raw test set contains 3,174 structure/cost credit cases, 7,002
preference-dependent credit cases, 4,039 accepted repairable-good cases, and
785 accepted non-repairable-good or already-satisfactory cases, summing to the
15,000-row test set. \racl-ValidatedGuard achieves perfect credit on both
structure/cost and preference-dependent groups. The right interpretation is not
that \racl{} uniquely discovers credit from labels. Rather, the repair ontology
is the identifying structure: \racl{} derives credit without explicit credit
labels, whereas label-only BlackBox does not; a black-box model wrapped with the
same repair search can also derive credit and plans.

\section{Limitations}

The main limitation is the known-repair-ontology assumption. \racl{} studies
repair-aware decision making when candidate repair operators, observable effects,
and cost/search spaces are specified. It does not discover arbitrary repair
operators from binary labels. Repair-library completeness matters: missing
repair templates remain a direct failure mode.

The raw DB1B-derived benchmark is also semi-synthetic. Its candidate
distribution comes from real ticket records, but ancillary repair fields,
contexts, labels, and credit categories are injected. The benchmark is a
controlled stress test on a real ticket distribution, not a natural airline
repair-choice dataset. Calibration guard selection is model selection, so we
report the validation-selected operating point alongside the fixed-guard
fallback. Finally, \racl{} is not optimized to be the best generic rank
classifier; AUROC and label accuracy are secondary to repair-aware decision
quality.

\section{Conclusion}

\racl{} reframes infeasibility as a repair-aware decision problem. Under a known
repair ontology, it generalizes no-repair feasibility-plus-preference semantics,
avoids the structural false-veto gap on repairable-good candidates, and produces
credit and repair-plan outputs without credit supervision. The DB1B-derived
stress test shows that validation-selected \racl{} can reduce false vetoes to
$10/4039$ while preserving structured explanations. The strongest repair-aware
black-box baseline obtains lower EDR but about $1064/4039$ false vetoes, so the
empirical picture is a precise operating trade-off rather than a blanket claim
of dominance. Future work should learn incomplete repair libraries, support
multi-step repair plans, integrate cost-sensitive repair-value trade-offs, and
evaluate on natural datasets where repair actions and user decisions are
observed rather than injected.

\bibliography{references}

@inproceedings{Kumar2020HASSLE,
  author = {Kumar, Mohit and Kolb, Samuel and Teso, Stefano and De Raedt, Luc},
  title = {Learning {MAX-SAT} from Contextual Examples for Combinatorial Optimisation},
  booktitle = {Proceedings of the Thirty-Fourth AAAI Conference on Artificial Intelligence},
  pages = {4493--4500},
  year = {2020},
  doi = {10.1609/aaai.v34i04.5877}
}

@article{Kumar2023ContextualMaxSAT,
  author = {Kumar, Mohit and Kolb, Samuel and Teso, Stefano and De Raedt, Luc},
  title = {Learning {MAX-SAT} from Contextual Examples for Combinatorial Optimisation},
  journal = {Artificial Intelligence},
  volume = {314},
  pages = {103794},
  year = {2023},
  doi = {10.1016/j.artint.2022.103794}
}

@inproceedings{Kolb2018SMT,
  author = {Kolb, Samuel and Teso, Stefano and Passerini, Andrea and De Raedt, Luc},
  title = {Learning {SMT(LRA)} Constraints Using {SMT} Solvers},
  booktitle = {Proceedings of the Twenty-Seventh International Joint Conference on Artificial Intelligence},
  pages = {2333--2340},
  year = {2018},
  doi = {10.24963/ijcai.2018/323}
}

@inproceedings{Bessiere2013QuAcq,
  author = {Bessiere, Christian and Coletta, Remi and Hebrard, Emmanuel and Katsirelos, George and Lazaar, Nadjib and Narodytska, Nina and Quimper, Claude-Guy and Walsh, Toby},
  title = {Constraint Acquisition via Partial Queries},
  booktitle = {Proceedings of the Twenty-Third International Joint Conference on Artificial Intelligence},
  pages = {475--481},
  year = {2013}
}

@article{Bessiere2017ConstraintAcquisition,
  author = {Bessiere, Christian and Koriche, Fr{\'e}d{\'e}ric and Lazaar, Nadjib and O'Sullivan, Barry},
  title = {Constraint Acquisition},
  journal = {Artificial Intelligence},
  volume = {244},
  pages = {315--342},
  year = {2017},
  doi = {10.1016/j.artint.2015.08.001}
}

@inproceedings{Tsochantaridis2004StructuredSVM,
  author = {Tsochantaridis, Ioannis and Hofmann, Thomas and Joachims, Thorsten and Altun, Yasemin},
  title = {Support Vector Machine Learning for Interdependent and Structured Output Spaces},
  booktitle = {Proceedings of the Twenty-First International Conference on Machine Learning},
  pages = {104--112},
  year = {2004},
  doi = {10.1145/1015330.1015341}
}

@inproceedings{Xu2018SemanticLoss,
  author = {Xu, Jingyi and Zhang, Zilu and Friedman, Tal and Liang, Yitao and Van den Broeck, Guy},
  title = {A Semantic Loss Function for Deep Learning with Symbolic Knowledge},
  booktitle = {Proceedings of the Thirty-Fifth International Conference on Machine Learning},
  series = {Proceedings of Machine Learning Research},
  volume = {80},
  pages = {5502--5511},
  publisher = {PMLR},
  year = {2018}
}

@inproceedings{Felfernig2008ConstraintRecommenders,
  author = {Felfernig, Alexander and Burke, Robin},
  title = {Constraint-Based Recommender Systems: Technologies and Research Issues},
  booktitle = {Proceedings of the 10th International Conference on Electronic Commerce},
  pages = {17--26},
  publisher = {ACM},
  year = {2008},
  doi = {10.1145/1409540.1409544}
}

@article{Felfernig2013RepairScoringRules,
  author = {Felfernig, Alexander and Schippel, Stefan and Leitner, Gerhard and Reinfrank, Florian and Isak, Klaus and Mandl, Monika and Blazek, Paul and Ninaus, Gerald},
  title = {Automated Repair of Scoring Rules in Constraint-Based Recommender Systems},
  journal = {AI Communications},
  volume = {26},
  number = {1},
  pages = {15--27},
  year = {2013},
  doi = {10.3233/AIC-120543}
}

@article{Boutilier2004CPNets,
  author = {Boutilier, Craig and Brafman, Ronen I. and Domshlak, Carmel and Hoos, Holger H. and Poole, David},
  title = {{CP}-Nets: A Tool for Representing and Reasoning with Conditional Ceteris Paribus Preference Statements},
  journal = {Journal of Artificial Intelligence Research},
  volume = {21},
  pages = {135--191},
  year = {2004},
  doi = {10.1613/jair.1234}
}

@article{Domshlak2011PreferencesAI,
  author = {Domshlak, Carmel and H{\"u}llermeier, Eyke and Kaci, Souhila and Prade, Henri},
  title = {Preferences in {AI}: An Overview},
  journal = {Artificial Intelligence},
  volume = {175},
  number = {7--8},
  pages = {1037--1052},
  year = {2011},
  doi = {10.1016/j.artint.2011.03.004}
}

@article{Wachter2018Counterfactual,
  author = {Wachter, Sandra and Mittelstadt, Brent and Russell, Chris},
  title = {Counterfactual Explanations without Opening the Black Box: Automated Decisions and the {GDPR}},
  journal = {Harvard Journal of Law and Technology},
  volume = {31},
  number = {2},
  pages = {841--887},
  year = {2018},
  doi = {10.2139/ssrn.3063289}
}

@inproceedings{Ustun2019Recourse,
  author = {Ustun, Berk and Spangher, Alexander and Liu, Yang},
  title = {Actionable Recourse in Linear Classification},
  booktitle = {Proceedings of the Conference on Fairness, Accountability, and Transparency},
  pages = {10--19},
  publisher = {ACM},
  year = {2019},
  doi = {10.1145/3287560.3287566}
}

@inproceedings{Karimi2021RecourseInterventions,
  author = {Karimi, Amir-Hossein and Sch{\"o}lkopf, Bernhard and Valera, Isabel},
  title = {Algorithmic Recourse: From Counterfactual Explanations to Interventions},
  booktitle = {Proceedings of the 2021 ACM Conference on Fairness, Accountability, and Transparency},
  pages = {353--362},
  publisher = {ACM},
  year = {2021},
  doi = {10.1145/3442188.3445899}
}

@inproceedings{Guo2017Calibration,
  author = {Guo, Chuan and Pleiss, Geoff and Sun, Yu and Weinberger, Kilian Q.},
  title = {On Calibration of Modern Neural Networks},
  booktitle = {Proceedings of the 34th International Conference on Machine Learning},
  series = {Proceedings of Machine Learning Research},
  volume = {70},
  pages = {1321--1330},
  publisher = {PMLR},
  year = {2017}
}

@inproceedings{Zadrozny2002Calibration,
  author = {Zadrozny, Bianca and Elkan, Charles},
  title = {Transforming Classifier Scores into Accurate Multiclass Probability Estimates},
  booktitle = {Proceedings of the Eighth ACM SIGKDD International Conference on Knowledge Discovery and Data Mining},
  pages = {694--699},
  publisher = {ACM},
  year = {2002},
  doi = {10.1145/775047.775151}
}

@incollection{Platt1999ProbabilisticOutputs,
  author = {Platt, John C.},
  title = {Probabilistic Outputs for Support Vector Machines and Comparisons to Regularized Likelihood Methods},
  booktitle = {Advances in Large Margin Classifiers},
  editor = {Smola, Alexander J. and Bartlett, Peter and Sch{\"o}lkopf, Bernhard and Schuurmans, Dale},
  pages = {61--74},
  publisher = {MIT Press},
  year = {1999}
}

@book{Vapnik1998SLT,
  author = {Vapnik, Vladimir N.},
  title = {Statistical Learning Theory},
  publisher = {Wiley},
  address = {New York},
  year = {1998}
}

@book{Anthony1999NNLearning,
  author = {Anthony, Martin and Bartlett, Peter L.},
  title = {Neural Network Learning: Theoretical Foundations},
  publisher = {Cambridge University Press},
  address = {Cambridge},
  year = {1999}
}

@book{ShalevShwartz2014UML,
  author = {Shalev-Shwartz, Shai and Ben-David, Shai},
  title = {Understanding Machine Learning: From Theory to Algorithms},
  publisher = {Cambridge University Press},
  address = {Cambridge},
  year = {2014}
}

@article{Friedman2001GradientBoosting,
  author = {Friedman, Jerome H.},
  title = {Greedy Function Approximation: A Gradient Boosting Machine},
  journal = {The Annals of Statistics},
  volume = {29},
  number = {5},
  pages = {1189--1232},
  year = {2001},
  doi = {10.1214/aos/1013203451}
}

@article{Pedregosa2011ScikitLearn,
  author = {Pedregosa, Fabian and Varoquaux, Ga{\"e}l and Gramfort, Alexandre and Michel, Vincent and Thirion, Bertrand and Grisel, Olivier and Blondel, Mathieu and Prettenhofer, Peter and Weiss, Ron and Dubourg, Vincent and Vanderplas, Jake and Passos, Alexandre and Cournapeau, David and Brucher, Matthieu and Perrot, Matthieu and Duchesnay, {\'E}douard},
  title = {Scikit-learn: Machine Learning in Python},
  journal = {Journal of Machine Learning Research},
  volume = {12},
  pages = {2825--2830},
  year = {2011}
}

@article{Fawcett2006ROC,
  author = {Fawcett, Tom},
  title = {An Introduction to {ROC} Analysis},
  journal = {Pattern Recognition Letters},
  volume = {27},
  number = {8},
  pages = {861--874},
  year = {2006},
  doi = {10.1016/j.patrec.2005.10.010}
}

@inproceedings{Davis2006PRROC,
  author = {Davis, Jesse and Goadrich, Mark},
  title = {The Relationship between Precision-Recall and {ROC} Curves},
  booktitle = {Proceedings of the 23rd International Conference on Machine Learning},
  pages = {233--240},
  publisher = {ACM},
  year = {2006},
  doi = {10.1145/1143844.1143874}
}

@book{Rossi2006CPHandbook,
  editor = {Rossi, Francesca and van Beek, Peter and Walsh, Toby},
  title = {Handbook of Constraint Programming},
  publisher = {Elsevier},
  year = {2006}
}

@article{Freuder1992PartialCSP,
  author = {Freuder, Eugene C. and Wallace, Richard J.},
  title = {Partial Constraint Satisfaction},
  journal = {Artificial Intelligence},
  volume = {58},
  number = {1--3},
  pages = {21--70},
  year = {1992},
  doi = {10.1016/0004-3702(92)90004-H}
}

@article{Bistarelli1997SemiringCSP,
  author = {Bistarelli, Stefano and Montanari, Ugo and Rossi, Francesca},
  title = {Semiring-Based Constraint Satisfaction and Optimization},
  journal = {Journal of the ACM},
  volume = {44},
  number = {2},
  pages = {201--236},
  year = {1997},
  doi = {10.1145/256303.256306}
}

@incollection{Meseguer2006SoftConstraints,
  author = {Meseguer, Pedro and Rossi, Francesca and Schiex, Thomas},
  title = {Soft Constraints},
  booktitle = {Handbook of Constraint Programming},
  editor = {Rossi, Francesca and van Beek, Peter and Walsh, Toby},
  pages = {281--328},
  publisher = {Elsevier},
  year = {2006}
}

@book{Biere2009SATHandbook,
  editor = {Biere, Armin and Heule, Marijn and van Maaren, Hans and Walsh, Toby},
  title = {Handbook of Satisfiability},
  publisher = {IOS Press},
  address = {Amsterdam},
  year = {2009}
}

@book{Gebser2012ASP,
  author = {Gebser, Martin and Kaminski, Roland and Kaufmann, Benjamin and Schaub, Torsten},
  title = {Answer Set Solving in Practice},
  publisher = {Morgan and Claypool},
  year = {2012},
  doi = {10.2200/S00457ED1V01Y201211AIM019}
}

@misc{BTSDB1B,
  author = {{Bureau of Transportation Statistics}},
  title = {Airline Origin and Destination Survey ({DB1B})},
  howpublished = {\url{https://transtats.bts.gov/DatabaseInfo.asp?QO_VQ=EFI}},
  note = {Accessed: 2026-05-29},
  year = {2026}
}

@misc{KaggleExpedia,
  author = {{Kaggle}},
  title = {Expedia Hotel Recommendations},
  howpublished = {\url{https://www.kaggle.com/competitions/expedia-hotel-recommendations}},
  note = {Accessed: 2026-05-29},
  year = {2016}
}

\end{document}